\documentclass[runningheads]{llncs}

\usepackage{eccv}

\usepackage{eccvabbrv}

\usepackage{graphicx}
\usepackage{booktabs}
\usepackage{multirow}

\usepackage[accsupp]{axessibility}  

\usepackage{hyperref}

\usepackage{orcidlink}

\usepackage{bbm}

\usepackage[normalem]{ulem}
\usepackage[table,xcdraw]{xcolor}
\newcommand{\best}[1]{\cellcolor{ForestGreen!20}$\mathbf{{#1}}$}
\newcommand{\second}[1]{\cellcolor{LimeGreen!20}\uline{{$#1$}}}
\newcommand{\third}[1]{\cellcolor{yellow!20}$#1$}

\DeclareRobustCommand{\bestcap}[1]{\colorbox{ForestGreen!20}{\textbf{#1}}}
\DeclareRobustCommand{\secondcap}[1]{\colorbox{LimeGreen!20}{\uline{#1}}}
\DeclareRobustCommand{\thirdcap}[1]{\colorbox{yellow!20}{#1}}

\usepackage{wrapfig}

\begin{document}

\title{Controllable Egocentric Video Generation via Occlusion-Aware Sparse 3D Hand Joints}

\titlerunning{Controllable Egocentric Video Gen. via Occlusion-Aware Sparse 3D Joints}

\author{
    Chenyangguang Zhang$^{1}$\thanks{Equal contribution.} \and
    Botao Ye$^{1,2\star}$ \and
    Boqi Chen$^{1,2\star}$ \and
    \mbox{Alexandros Delitzas$^{1,3}$} \and
    Fangjinhua Wang$^{1}$ \and
    Marc Pollefeys$^{1,4}$ \and
    Xi Wang$^{1}$
}

\authorrunning{C. Zhang et al.}

\institute{
    $^1$ETH Zurich \quad 
    $^2$ETH AI Center \quad
    $^3$MPI for Informatics \quad $^4$Microsoft
}

\maketitle

\let\thefootnote\relax\footnotetext{Project page: \url{https://zhangcyg.github.io/handcontrolvideo/}}

\vspace{-4mm}
\begin{abstract}
Controllable video generation for complex hand-object interactions is a critical step toward building visual world models.
However, existing methods often struggle to achieve fine-grained, 3D-consistent hand articulation in generated videos.
By relying on dense 2D trajectories or implicit pose representations, they collapse crucial geometric structures into spatially ambiguous signals, leading to severe motion inconsistencies and hallucinated artifacts under egocentric occlusions.
To address this, we propose leveraging sparse 3D hand joints as explicit control signals with three key advantages: explicit geometry to resolve occlusions, an intuitive interface for interactive editing, and cross-embodiment generalization to robotic hands.
Built upon this, our efficient control module extracts occlusion-aware features from the source reference frame by penalizing unreliable visual features from hidden joints, and employs a 3D-based weighting mechanism to handle dynamically occluded target joints during motion propagation.
Meanwhile, it directly injects 3D geometric embeddings into the latent space to enforce structural consistency.
To facilitate robust training and evaluation, we develop an automated annotation pipeline, yielding 1M high-quality egocentric video clips paired with precise hand trajectories.
Experiments demonstrate that our approach outperforms state-of-the-art baselines, generating high-fidelity egocentric videos with realistic hand-object interactions.

\end{abstract}

\section{Introduction}

\begin{figure}[ht]
  \centering
\includegraphics[width=0.95\linewidth]{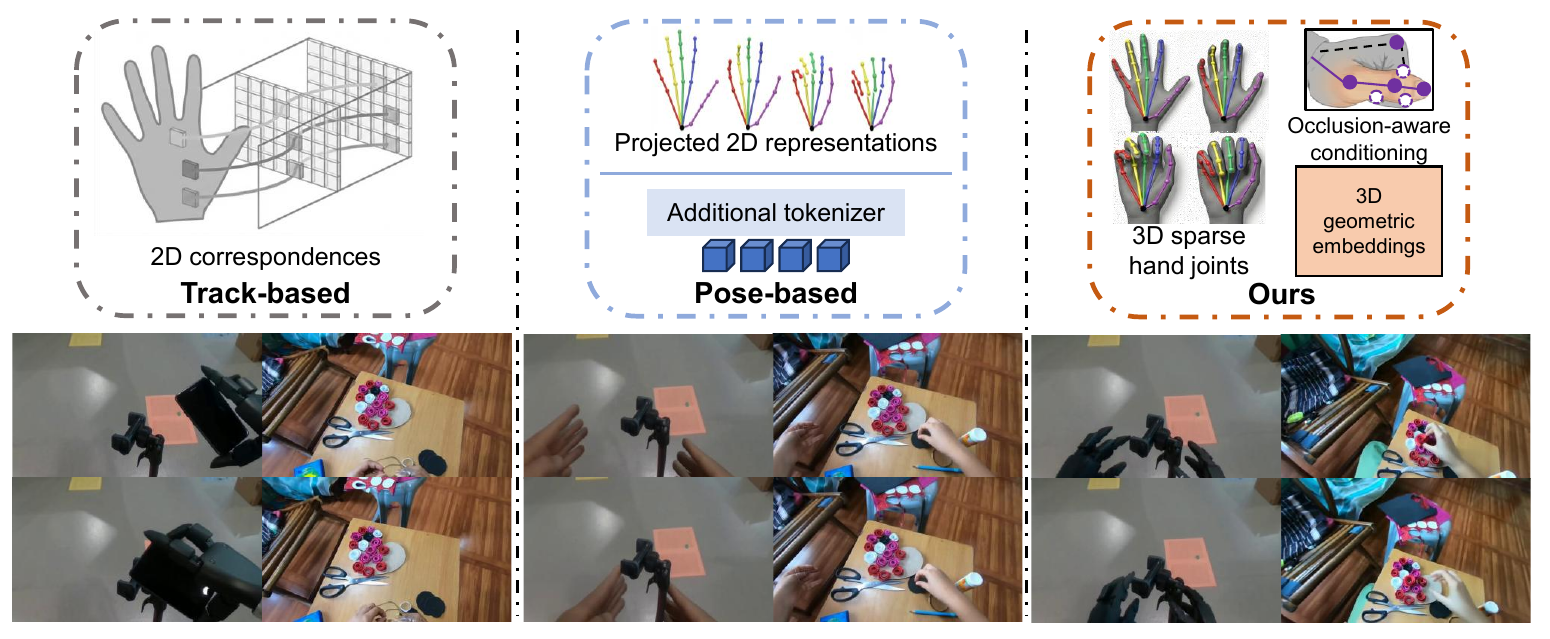}
   \vspace{-3mm}
   \caption{\textbf{Comparison to existing controllable video generation methods.} 
   In contrast to existing track- and pose-based methods, we leverage sparse 3D hand joints as a direct control signal, unlocked by our occlusion-aware conditioning and 3D geometric embeddings. 
   As a result, our method generates crisp, high-fidelity hands that align with complex motions even under heavy occlusion across different embodiments.}
   \vspace{-7mm}
   \label{fig:teaser}
\end{figure}

The recent evolution of generative models is pushing video synthesis beyond mere content creation, laying the foundation for visual world models capable of simulating complex physical realities~\cite{wan2025wan,ali2025world,blattmann2023stable,jin2024pyramidal,wu2025hunyuanvideo}. 
To realize this potential, research has increasingly shifted towards motion-controlled video generation, with the aim of giving precise authority over action dynamics \cite{wang2024motionctrl,geng2025motion,chu2025wan, shin2025motionstream, lee2025generative}. 
Such controllability is particularly critical in egocentric scenarios, where fine-grained hand-object interactions determine perceptual realism.
In such case, the model must simultaneously preserve high-frequency articulation details, maintain geometric plausibility under frequent occlusion, and follow action trajectories with strict spatial consistency. 

However, existing control paradigms struggle to satisfy these requirements.
Track-based methods \cite{chu2025wan,geng2025motionprompting,xiao2025trajectoryattention,xing2025motioncanvas,shin2025motionstream} treat motion flows as isolated 2D point trajectories, discarding the rigid structural integrity of the hand and inherently lacking depth awareness, resulting in unrealistic motion details when facing complex interactions. 
Pose-based methods \cite{pallotta2025egocontrol, li2025mask2iv} represent hand motion using implicit pose signals and generally follow two strategies. 
One encodes poses into compressed, low-frequency latent manifolds~\cite{pallotta2025egocontrol,tu2025playerone}, which smooths fine-grained articulation details and thus weakens the high-frequency spatial cues required by generators, often producing temporally averaged or floaty motion artifacts \cite{zhang2023adding,zhang2025vhoi,li2025mask2iv,han2023controllable}.
The other projects poses into 2D skeletal maps or masks \cite{li2025mask2iv,Wan2_1_Fun_Control2025}. 
Although interpretable, such projections collapse 3D geometry into ambiguous 2D signals. 
When fingers intersect or occlude each other, overlapping structures create visual clutter and ambiguous ordering cues that hinder precise control (Figure~\ref{fig:teaser}).
Moreover, the implicit and latent nature of the pose representation precludes intuitive interactive editing in user interfaces.

Our insight is that sparse 3D hand joints offer a more suitable control signal for egocentric video generation by providing:
(1) explicit 3D geometry to resolve occlusion ambiguities, 
(2) a sparse, intuitive interface for interactive user edit,
(3) an embodiment-agnostic representation that naturally supports cross-embodiment generalization.
Specifically, unlike dense skeletal maps or latent pose codes that embed rigid human-specific kinematic assumptions, sparse 3D joints represent articulations as unconstrained spatial coordinates. 
This flexibility ensures our framework not only enhances the fidelity of human egocentric videos but seamlessly transfers to robotic dexterous manipulation \cite{qin2022dexmv,xin2026analyzing} without requiring complex domain adaptation (See Supplementary Material).

However, directly injecting 3D joints into an egocentric image-to-video generation model is non-trivial and introduces two fundamental challenges.
\textbf{First, occlusion ambiguity.} Egocentric interactions involve severe and dynamic mutual occlusions between fingers and objects. The model must explicitly reason about visibility states to avoid feature contamination and trajectory conflicts. This challenge comes in two forms. 
(1) \textit{Source occlusion}: in the reference frame, occluded joints should not contribute unreliable visual features. Otherwise, hidden articulations may inherit incorrect textures, leading to hallucinations when later revealed. 
(2) \textit{Target occlusion}: during motion propagation, foreground fingers frequently cross over background ones. Without 3D-aware conditioning, features may be incorrectly assigned across layers, producing anatomical artifacts such as merged or intersecting fingers.
\textbf{Second, preservation of 3D structure.} 
A naive conditioning strategy that projects 3D joints into 2D feature space discards depth and structural priors, undermining the key advantage of 3D control. 
To achieve high-fidelity generation, semantic identity and geometric relationships must be incorporated directly into the generation process.

To address these challenges, we propose a novel occlusion-aware, geometry-preserving control framework for egocentric image-to-video generation (Figure~\ref{fig:teaser}).
First, we construct occlusion-aware motion condition features by explicitly modeling both source and target occlusion.
In the source reference frame, we sample latent VAE features by aggregating local visual contexts around each joint, while filtering out unreliable signals via an occlusion penalty strategy.
Then, these refined features are propagated to target frames along the 3D hand joint trajectories. 
To handle dynamic target occlusions during motion, we introduce a 3D-based weighting mechanism that adaptively adjusts feature contributions of the joints, ensuring robust motion feature conditioning during severe mutual occlusion. 
Finally, to fully exploit the semantic and geometric richness of sparse hand joints in the generation process, we construct 3D geometric embeddings incorporating both 3D coordinate information and learnable semantic indices.
Despite explicitly modeling occlusion and 3D geometry, the entire control module remains lightweight and minimally perturbs the pretrained latent space, allowing efficient integration with modern video generation backbones.

Furthermore, we observe a lack of large-scale benchmarks for training and evaluating controllable egocentric video generation models with accurate 3D hand trajectories. 
To bridge this gap, we develop an automated annotation and filtering pipeline. 
Using the Ego4D dataset \cite{grauman2022ego4d}, we produce 1M high-quality video clips with accurate hand trajectories. 
Extensive experiments on the Ego4D and EgoDex \cite{hoque2025egodex} datasets demonstrate that our method outperforms existing approaches, achieving superior accuracy in controlling complex egocentric motions under heavy occlusion. 
Moreover, our framework uniquely enables interactive, fine-grained micro-manipulation, allowing users to control articulations down to the isolated movement of a single joint.

\section{Related Works}

\noindent\textbf{Video Generation Models.}
Early advances in video generation were primarily driven by Latent Diffusion Models~\cite{ldm} (\textit{e.g.}, 3D U-Net~\cite{blattmann2023stable}) to maintain spatial-temporal coherence. 
More recently, the field has undergone a paradigm shift toward Diffusion Transformers (DiTs)~\cite{dit}, which leverage the scalability of transformer backbones to handle long-range temporal dependencies. 
State-of-the-art open-source models such as WAN~\cite{wan2025wan} and HunyuanVideo~\cite{hunyuanvideo} have significantly narrowed the quality gap between open and proprietary systems with exceptional photorealistic detail. 
Despite their impressive generative priors, these DiT-based models lack the mechanisms for precise, fine-grained control over local dynamics. 
Our work provides concise, 3D-aware control using sparse hand joints as explicit signals to guide complex egocentric video generation. 

\noindent\textbf{Motion Control in Video Generation.}
Early explorations into motion-controlled video generation focused on training-free methods~\cite{namekata2024sgi2v,qiu2024freetraj,ma2024trailblazer,jain2024peekaboo}, which optimize input noisy latents or manipulate attention mechanisms to achieve zero-shot guidance. 
Subsequently, finetuning-based approaches have become the dominant paradigm~\cite{wang2023videocomposer,burgert2025gowiththeflow,wang2024motionctrl,li2025magicmotion,zhang2025tora,li2025imageconductor,fu2024trajmaster,wang2025cinemaster,zhou2025trackgo,shi2024motioni2v,wang2024replaceanyone,xia2025dreamve,geng2025motionprompting,hassan2025gem}, using auxiliary adapters to inject various motion signals into the base model. 
Among various control modalities, point tracks have emerged as a unified representation for guiding general motions~\cite{geng2025motionprompting,shi2024motioni2v,wang2025ati,wang2024motionctrl,wu2024draganything,xiao2025trajectoryattention,xing2025motioncanvas,zhang2025tora}. 
Most recently, state-of-the-art methods such as Wan-Move~\cite{chu2025wan} and MotionStream~\cite{shin2025motionstream} have advanced this paradigm by injecting 2D point trajectories directly into the latent space to provide motion context. 
Despite their general success, 2D-track-based methods discard the structural integrity and 3D-awareness essential for egocentric interactions, often failing to resolve severe occlusions and causing structural hallucinations. 
To overcome this, our method leverages sparse 3D hand joints, explicitly tackling the inherent challenges of 3D-to-2D information loss through occlusion-aware conditioning and 3D geometric embeddings.

\noindent\textbf{Egocentric World Models.}
Rapid advances in video generation have catalyzed the development of egocentric world models that simulate first-person dynamics to facilitate applications in Virtual Reality (VR) and Robotics.
A prominent line of research focuses on synthesizing egocentric content that depicts a variety of controlled human behaviors, by conditioning generative priors on modalities of human poses, trajectories, and action signals \cite{tu2025playerone,zhang2025vhoi,pallotta2025egocontrol,bagchi2026walk,wang2026hand2world,wang2025precise,wang2025humandreamer,zhang2025egocentric,hassan2025gem}. 
Building upon these human-centric simulations, a parallel body of work seeks to bridge the embodiment gap for the robotics community \cite{hafner2025training,jiang2025rynnvla,lykov2025physicalagent,shi2026egohumanoid,bi2025motus,gao2026dreamdojo,ni2024recondreamer,mendonca2023structured,fung2025embodied}. 
These approaches translate human behavioral priors into embodied robots.
While existing methods successfully capture holistic dynamics, they often lack the fine-grained kinematic fidelity required to simulate dexterous interactions. 
Our method integrates precise 3D control into generative world models via sparse joints, enabling high-fidelity synthesis for VR rendering and robotic learning.
\section{Method}

\begin{figure}[ht]
  \centering
   \includegraphics[width=0.95\linewidth]{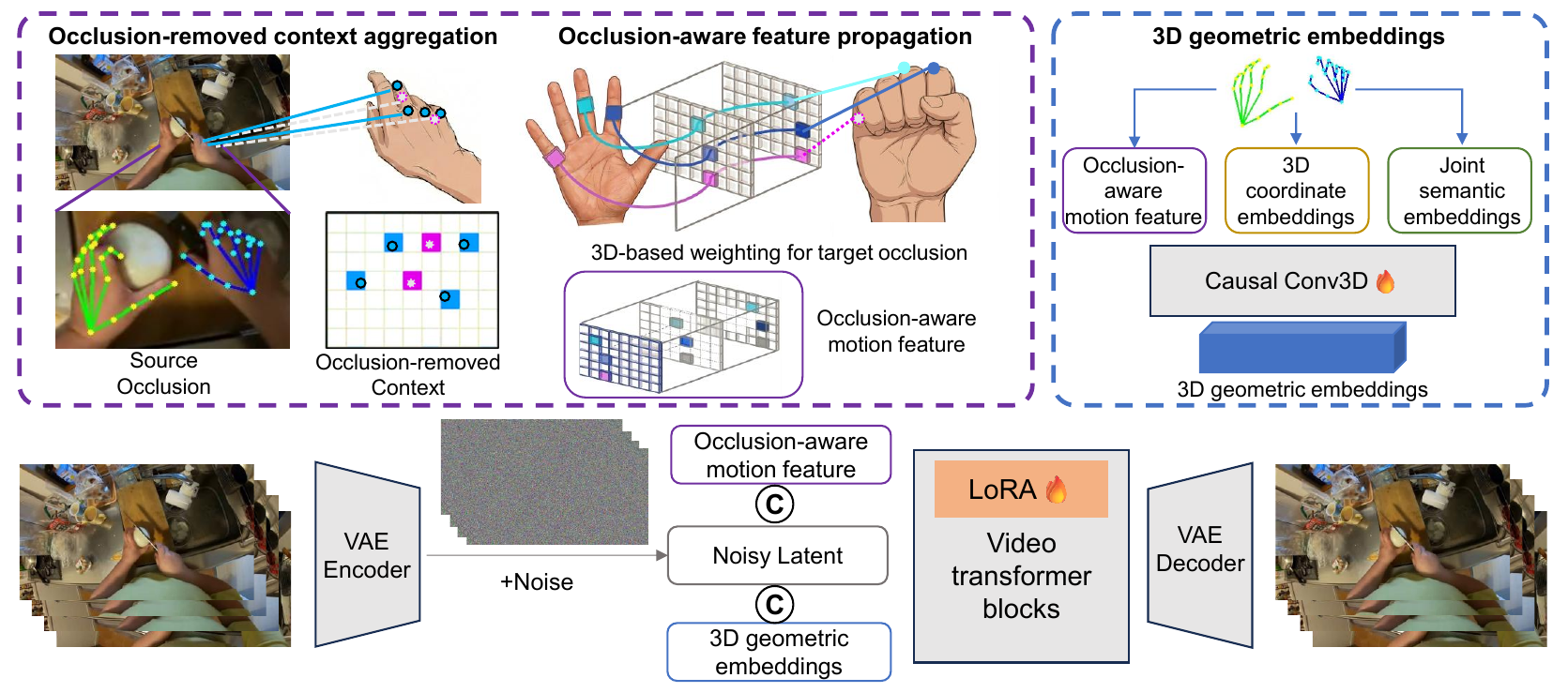}
   \vspace{-3mm}
   \caption{\textbf{Method overview.} Our framework uses sparse 3D hand joints to represent motions by constructing two embedding streams.
   The occlusion-aware motion feature 
   is yielded by first penalizing occluded regions to extract reliable context from the source frame, and then propagating it with modulating 3D-aware feature weights to handle target occlusion. 
   The 3D geometric embedding is formed by processing this motion feature along with 3D joint coordinates and semantic embeddings through a Causal Conv3D block.
   Finally, both embeddings are concatenated with the noisy latent and fed into LoRA-adapted DiT blocks.}
   \vspace{-7mm}
   \label{fig:pipeline}
\end{figure}

\subsection{Preliminaries: Video Generation Model}

Our method builds upon the WAN framework~\cite{wan2025wan}, which formulates video generation as a continuous-time conditional flow matching problem.
Flow matching models the generative process as a deterministic Ordinary Differential Equation (ODE) that transports samples from a simple prior distribution toward the data distribution.
Let $x_1 \sim p_{\text{video}}$ denote the latent representation of a real video encoded by a 3D-VAE, and let $x_0 \sim \mathcal{N}(0, I)$ be Gaussian noise sampled from the prior distribution.
A conditional probability path is defined by linear interpolation between $x_0$ and $x_1$: $x_t = (1 - t)x_0 + t x_1, t \in [0,1]$.
The time derivative of this path is constant and is given by $\frac{\mathrm{d}x_t}{\mathrm{d}t} = x_1 - x_0.$

A time-conditioned neural network $v_\theta(x_t, t, c)$, parameterized as a DiT conditioned on multimodal context $c$ (\textit{e.g.}, texts, reference images, or control signals), is trained to approximate this target velocity field with the objective:
\begin{equation}
\mathcal{L} 
= \mathbb{E}_{t \sim \mathcal{U}(0,1),\, x_0,\, x_1,\, c}
\left[
\left\| v_\theta(x_t, t, c) - (x_1 - x_0) \right\|^2
\right].
\label{eq:flow_loss}
\end{equation}

At inference time, video samples are generated by solving the learned ODE $\frac{\mathrm{d}x_t}{\mathrm{d}t} = v_\theta(x_t, t, c)$, from $t=0$ to $t=1$ using a numerical ODE solver, thereby transporting the prior sample to the video data manifold.

The standard WAN framework conditions the velocity field $v_\theta$ in a global context, which is insufficient to define precise dexterous manipulations.
Our work focuses on fine-grained motion control for egocentric video generation by introducing sparse 3D hand joints $\mathbf{J} \in \mathbb{R}^{F \times N \times 3}$ as an explicit motion guide. 

\subsection{Occlusion-Aware 3D Hand Motion Control}

Our framework (Figure \ref{fig:pipeline}) integrates sparse 3D hand joints into the image-to-video generation process by addressing two specific challenges of egocentric interactions. 
First, to resolve the ambiguity of mutual occlusions, we devise an occlusion-aware motion conditioning strategy. 
This involves a two-stage process: 
(1) Occlusion-removed context aggregation in the source reference frame to prevent feature contamination from occluding parts, and (2) Occlusion-aware feature propagation to modulate feature influence during target frame generation.
Second, to fully leverage the spatial richness of the 3D joints, we align 3D geometric embeddings directly into the video latent space, encoding both articulation semantics and 3D consistency.

\noindent\textbf{Occlusion-Removed Context Aggregation.}
To guide the generation of subsequent frames, we first aim to extract visual context features from the VAE latent of the source image corresponding to each hand joint. 
However, two obstacles prevent simple feature sampling, \ie, \textit{sparsity} and \textit{source occlusion}.

Unlike track-based methods~\cite{chu2025wan,shin2025motionstream} that rely on dense point grids for better performance \cite{chu2025wan}, hand joints are inherently sparse ($N \le 42$). 
Relying solely on single-pixel features at projected coordinates captures insufficient context for complex hand-object interactions.
To address this, we aggregate local features by modeling the spatial influence of each joint.
Given the set of 3D hand joints $\mathcal{J}_{src} = \{\mathbf{J}_i\}_{i=1}^N$ in the source camera frame, we project them to 2D coordinates and disparity (inverse depth) maps $\mathbf{u}_i, d_i = \Pi(\mathbf{J}_i, \mathbf{K})$ using camera intrinsic $\mathbf{K}$. 
We then generate a localized Gaussian heatmap $\mathbf{M}_i(\mathbf{x}) = \exp \left( -\frac{\| \mathbf{x} - \mathbf{u}_i \|_2^2}{2\sigma^2} \right) \in \mathbb{R}^{H \times W}$ at a grid location $\mathbf{x} \in \mathbb{R}^2$ for each joint $i$, where $\sigma$ controls the spatial bandwidth, defining the receptive field for context aggregation.

While this Gaussian pooling captures context, it introduces a risk of feature contamination under self-occlusion. 
For instance, if a thumb occludes an index finger in the source view, naive aggregation around the index finger's projected position would erroneously sample the thumb's texture. Using this contaminated feature to generate the index finger in later frames would lead to severe hallucinations.
To resolve this, we define a pairwise occlusion penalty $P_{i \leftarrow j}$ representing the likelihood that joint $j$ occludes joint $i$:
\begin{equation}
    P_{i \leftarrow j} = \underbrace{\exp \left( -\frac{\| \mathbf{u}_i - \mathbf{u}_j \|_2^2}{2\tau^2} \right)}_{\text{Spatial Overlap}} \cdot \underbrace{\sigma \left( \gamma (d_j - d_i) \right)}_{\text{Depth Ordering}},
\end{equation}
where $\sigma(\cdot)$ is the sigmoid function. The term $\sigma(\gamma(d_j - d_i))$ approaches 1 when the joint $j$ is significantly closer to the camera than the joint $i$ ($d_j > d_i$), indicating a high probability of occlusion.

Finally, we extract the purified context feature $\mathbf{f}_i$ from the source latent map $\mathcal{Z}_0$. 
We apply a gated average pooling of features by:
\begin{equation}
    \mathbf{f}_i = (1 - \max_{j \neq i} (P_{i \leftarrow j})) \cdot \sum_{\mathbf{x}} \left( \frac{\mathbf{M}_i(\mathbf{x})}{\sum_{\mathbf{x}'} \mathbf{M}_i(\mathbf{x}') + \epsilon} \cdot \mathcal{Z}_0(\mathbf{x}) \right).
\end{equation}
This operation effectively filters the signal: if a joint is occluded (with visibility $1 - \max_{j \neq i} \approx 0$), its contribution to the feature is suppressed, preventing the propagation of erroneous texture information.

\noindent\textbf{Occlusion-Aware Feature Propagation.}
Once the clean source features $\mathbf{f}_i$ are extracted, the next challenge is how to propagate them to the target frames $t \in \{1, \dots, T\}$ along the trajectories of the 3D hand joints, to provide the generation process with accurate motion guidance. 
A naive summation of feature maps at target locations \cite{chu2025wan} fails due to \textit{target occlusion}: 
when a foreground finger crosses a background finger, their projections overlap. 
Without explicit 3D handling, features would mix erroneously, or background features might obscure the foreground.
To resolve this trajectory conflict, we propose a 3D-based weighting mechanism. 
It acts as a differentiable Z-buffer, dynamically assigning higher importance to joints closer to the camera during feature propagation.

For each frame $t$, we generate target Gaussian heatmaps $\mathbf{M}_{i,t} \in \mathbb{R}^{H \times W}$ centered at the projected joint locations $\mathbf{u}_{i,t}$, and retrieve their corresponding disparity values $d_{i,t}$. We compute a dense attention weight map $\mathbf{A}_{i,t}(\mathbf{x})$ for each joint $i$ at pixel $\mathbf{x}$, modulating spatial heatmap intensity with depth information:
\begin{equation}
    \mathbf{A}_{i,t}(\mathbf{x}) = \text{softmax}_i \left( \log(\mathbf{M}_{i,t}(\mathbf{x}) + \epsilon) + \lambda \cdot d_{i,t} \right),
\end{equation}
where $\lambda$ is a learnable scaling factor that controls the sharpness of the depth prioritization. 
The logarithmic term ensures that the base influence is determined by spatial proximity, while the linear disparity term boosts the logits for foreground joints. Consequently, at pixels where multiple joints overlap, the softmax allocates the majority of the weight to the joint closest to the camera.

The final motion-conditioned feature map $\mathcal{F}_{motion}(\mathbf{x}, t)$ is obtained by the weighted summation of the source features:
\begin{equation}
    \mathcal{F}_{motion}(\mathbf{x}, t) = \underbrace{\left( \sum_{i=1}^N \mathbf{A}_{i,t}(\mathbf{x}) \cdot \mathbf{f}_i \right)}_{\text{Normalized Feature}} \cdot \underbrace{\left( \sum_{j=1}^N \mathbf{M}_{j,t}(\mathbf{x}) \right)}_{\text{Total Opacity}}.
\end{equation}
The normalization ensures features are convex combinations of joints, while the final opacity term masks out background regions where no joints are present, to ensure the purified motion guidance.

Finally, we integrate these propagated features into the video generation process.
We first compress the motion feature volume $\mathcal{F}_{motion}$ in the time dimension for matching the VAE latent following \cite{chu2025wan}, then explicitly inject it into the image-to-video condition tensor $\mathbf{y}$ of the generation model. 
Specifically, for all subsequent frames $t > 0$, we populate the channels of $\mathbf{y}$ with our occlusion-aware motion features, while preserving the reference image features at $t=0$. 


\noindent\textbf{3D Geometric Embeddings.}
While the occlusion-aware propagation ensures that visual textures are correctly mapped to target locations, relying solely on propagated visual features is still insufficient. 
Visual features are inherently 2D representations: the 3D-to-2D projection inevitably compresses geometric information and obscures semantic identities of different articulations.
To compensate these, we introduce a secondary control stream that explicitly injects 3D information into the generation process offered by 3D spatial coordinates $(u, v, d)$, and distinct semantic definitions for each index of sparse 3D hand joints.

For each joint $i$ at frame $t$, we construct a high-dimensional embedding $\mathbf{z}_{i,t}$ that fuses its spatial location with its semantic identity. 
We employ sinusoidal positional encodings $\gamma(\cdot)$ to map the projected 3D coordinates (2D position $\mathbf{u}_{i,t}$ and disparity $d_{i,t}$) into a continuous frequency domain, and concatenate this with a learnable joint identity embedding $\mathbf{E}_{id}[i]$:
\begin{equation}
    \mathbf{z}_{i,t} = \phi \left( \left[ \gamma(\mathbf{u}_{i,t}, d_{i,t}) \mathbin{;} \mathbf{E}_{id}[i] \right] \right),
\end{equation}
where $\phi(\cdot)$ is a shallow MLP projector. 
This indicates that the model explicitly knows which part of the hand resides at which 3D depth.

To align these sparse joint embeddings with the video latent space, we splat them onto the spatial feature grid using the previously computed Gaussian heatmaps $\mathbf{M}_{i,t}$. 
This creates a dense geometric map $\mathcal{F}_{geo}(\mathbf{x}, t)$ at each pixel:
\begin{equation}
    \mathcal{F}_{geo}(\mathbf{x}, t) = \sum_{i=1}^N \mathbf{M}_{i,t}(\mathbf{x}) \cdot \mathbf{z}_{i,t}.
\end{equation}

We concatenate the geometric map $\mathcal{F}_{geo}$ and the motion cues $\mathcal{F}_{motion}$ to construct the final geometric embeddings.
To enforce temporal consistency, respect the causal nature of video generation, while matching the distribution of the VAE latent, we process this volume via a causal 3D convolutional head:
\begin{equation}
    \mathcal{C}_{geo} = \text{LayerNorm} \left( \text{Conv3D}_{causal} \left( [\mathcal{F}_{geo} \mathbin{;} \mathcal{F}_{motion}] \right) \right).
\end{equation}
The causal convolution applies asymmetric padding along the time dimension, ensuring that the geometric guidance for the current frame is aggregated by historical context without leaking information from future frames. 

To drive the video generation, we construct a comprehensive input for the video transformer. 
We concatenate the geometric control volume $\mathcal{C}_{geo}$ with the noisy video latent $\mathcal{X}$ and the visual condition tensor $\mathbf{y}$ (which contains the reference image features and motion-injected channels) along the channel dimension. 
Crucially, unlike computationally expensive ControlNet-like counterparts~\cite{zhang2023adding,pallotta2025egocontrol,Wan2_1_Fun_Control2025} that introduce significant overhead and risk destabilizing the pre-trained latent distribution, our strategy is lightweight ($\sim$20k parameters) while minimizing interference with the original latent space structure, allowing to achieve high-fidelity video quality and precise 3D control accuracy with negligible parameter cost.

\subsection{Training and Inference}
We build our framework upon the state-of-the-art image-to-video model WAN 2.1~\cite{wan2025wan}. 
To efficiently adapt the pre-trained backbone to egocentric motion control, we employ Low-Rank Adaptation (LoRA)~\cite{hu2022lora} with rank $r=64$ applied to the transformer layers. 
Other parameters introduced in the motion control module are trained from scratch. 
The entire framework is trained on our large-scale egocentric data (Section \ref{sec:data}) on 16 NVIDIA GH200 GPUs for $\sim$48 hours.

To ensure the framework's robustness against imperfect control signals, we implement a stochastic joint masking strategy. 
During training, we randomly mask $5\%$ of the input hand joints, setting all their coordinates to zero. 
Consequently, the model acquires the ability to plausibly generate missing articulations, ensuring temporal consistency when the upstream tracker suffers from intermittent failures or users provide sparse input during interactive editing.

During inference, the explicit nature of the 3D joint representation offers superior flexibility compared to prior methods. 
Our framework supports two distinct control modes: 
(1) Sequence Driven, where the video is driven by full trajectory sequences, and (2) Interactive Editing, where users intuitively define trajectories by dragging a single joint (or a subset of joints). 
In the latter mode, our model leverages its learned geometric priors to generate plausible articulation dynamics following the user-defined guidance (Section~\ref{sec:inter_vis}).
\section{Dataset Construction}\label{sec:data}

We adopt the Ego4D \cite{grauman2022ego4d} dataset as the large-scale video corpus for our egocentric video generation framework, which requires precise and temporally coherent hand joints to serve as control signals. 
Given the absence of 3D hand joint annotations in Ego4D, we propose an automated, scalable pipeline with high-fidelity hand pose extraction, spatiotemporal refinement, and quality assurance. 

\noindent\textbf{Hand Pose Extraction.}
To address severe motion blur, rapid camera movement, and frequent occlusions of unconstrained egocentric footage, we employ a decoupled frame-by-frame reconstruction paradigm. 
Initially, each video frame is processed by a YOLO-based detector \cite{redmon2018yolov3} to extract tight hand bounding boxes and categorical handedness signals (left or right). 
These localized image crops are subsequently input into WiLoR \cite{potamias2025wilor}, a state-of-the-art framework for in-the-wild 3D hand reconstruction, to predict MANO parameters \cite{romero2022embodied}, global root orientation, and camera translation. 

\noindent\textbf{Spatiotemporal Refinement.}
To synthesize temporally continuous control signals necessary for video generation, independent frame-wise predictions must be resolved into consistent tracks. 
This is particularly challenging in egocentric scenarios due to complex hand-object interactions and occlusions. 
We formulate the tracking objective as a bipartite matching problem in two primary topological slots, denoted $S \in \{L, R\}$.
In each frame $t$, we compute an assignment cost matrix $\mathbf{C}$ between current detections $h_i^{(t)}$ and established track slots $S_j$, integrating spatial continuity and semantic priors by:
\begin{equation}
    \mathbf{C}_{i,j} = \| \mathbf{T}_i^{(t)} - \mathbf{T}_j^{(t-1)} \|_2 + \lambda \mathbbm{1}(p_i^{(t)} \neq c_j),
\end{equation}
where $\mathbf{T}_i^{(t)}$ represents the 3D global translation of the detected hand, $\mathbf{T}_j^{(t-1)}$ is the last known spatial coordinate of the track $S_j$, $p_i^{(t)}$ is the localized handedness, $c_j$ is the nominal class of the slot, and $\lambda=0.05$ penalizes handedness mismatch.

To mitigate track flickering and identity inversions during close-proximity hand interactions, we introduce a temporal hysteresis threshold $\tau_{swap}$. 
A track swap is strictly prohibited unless the spatial cost improvement satisfies $\Delta \mathbf{C} > \tau_{swap}$. 
Furthermore, we incorporate a deterministic gap-reset mechanism:
if a tracked entity is occluded for a temporal window exceeding $\tau_{gap} = 10$ frames, the track history is flushed. 
This effectively prevents erroneous long-distance spatial snapping upon the re-emergence of the hand.

\begin{figure}[t]
    \centering
    \includegraphics[width=0.95\linewidth]{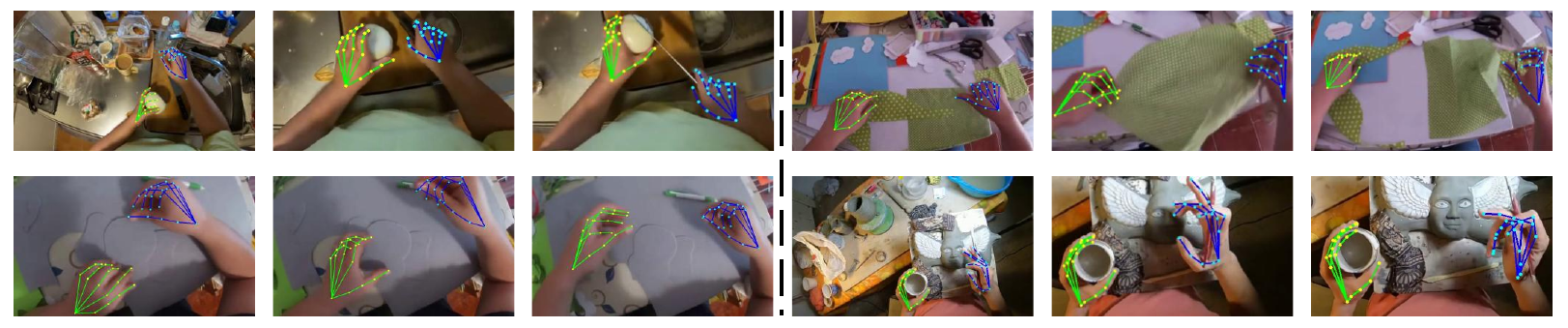}%
    \vspace{-3mm}
    \caption{\textbf{Qualitative results of our data annotations.} 
    The last two images are zoomed in for clear visualization of hand tracking accuracy.}
    \label{fig:data}
    \vspace{-7mm}
\end{figure}

\noindent\textbf{Quality Assurance.}
Generative models exhibit high sensitivity to stochastic noise in control conditions, necessitating rigorous quality assurance. 
We therefore filter the candidate sequences by discarding video clips where valid hand instances appear in fewer than 20\% of the frames, or valid MANO configurations fall below a 5\% density threshold. 
We further eliminate sequences where more than two hands are detected for over 25\% of the sequence duration.
After filtering, we obtain approximately 1 million high-quality video clips.
Figure~\ref{fig:data} presents qualitative examples of our pipeline's joint annotations projected into 2D image space. 
By leveraging this automated processing pipeline, we can efficiently generate large-scale, high-fidelity hand trajectory annotations with exceptional accuracy.

For high-quality tracks after filtering, the optimized MANO parameters are mapped to 3D joint coordinates via the forward kinematic function:
\begin{equation}
    \mathbf{J} = \mathcal{W}(\mathcal{M}(\boldsymbol{\theta}, \boldsymbol{\beta}), \mathbf{R}_{root}, \mathbf{T}),
\end{equation}
where $\mathcal{M}$ represents the MANO mesh topology defined by the pose vector $\boldsymbol{\theta}$ and the shape vector $\boldsymbol{\beta}$, which produces 3D joints $\mathbf{J} \in \mathbb{R}^{21 \times 3}$ each hand after the application of the global root rotation $\mathbf{R}_{root}$ and translation $\mathbf{T}$. 
All extracted right-hand joints are geometrically mirrored across the sagittal (X) axis, effectively homogenizing the generative conditioning space.

\section{Experiments}
\subsection{Experimental Setup}

\noindent\textbf{Evaluation Datasets.}
To comprehensively assess our method's generative capabilities and control accuracy, we construct evaluation splits from two distinct egocentric video benchmarks. 
For in-domain evaluation, we utilize the Ego4D dataset~\cite{grauman2022ego4d}, a massive-scale collection of various egocentric activities in daily life. 
To assess out-of-domain generalization, we employ the EgoDex dataset~\cite{hoque2025egodex}. 
This benchmark specifically focuses on diverse, dexterous tabletop manipulations that serve as critical priors for robotic visual-motor control.
In total, the evaluation suite consists of 40 unseen video sequences from Ego4D and 30 sequences from EgoDex. 
These test clips are randomly sampled to ensure a diverse representation of challenging, real-world interactions, encompassing scenarios such as kitchen activities, fine-grained tabletop manipulation, and dynamic tool usage.

\noindent\textbf{Metrics.}
To evaluate the fidelity of generated videos, we report standard video quality metrics, including FID \cite{heusel2017gans}, FVD \cite{unterthiner2019fvd}, PSNR, and SSIM \cite{wang2004image}. 
To assess the accuracy of the generated hand motions, we use WiLoR \cite{potamias2025wilor} to extract 3D hands from the predicted and ground truth videos in the camera coordinate and report the procrustes-aligned Mean Per-Joint Position Error (MPJPE).  
We further evaluate 3D hand shape accuracy using Mean Per-Vertex Position Error (MPVPE) over all hand mesh vertices.

\noindent\textbf{Baselines.}
We compare our method with representative approaches covering mask-based, pose-based, and track-based control paradigms.
Mask2IV~\cite{li2025mask2iv} adopts interaction masks as explicit control signals.
We evaluate its second stage only, where the video diffusion model is conditioned on per-frame interaction masks.
Two variants are reported: Mask2IV(H), which conditions on hand masks only, and Mask2IV(HO), which additionally incorporates object masks with ground-truth trajectories.
WAN-Fun~\cite{Wan2_1_Fun_Control2025}, built upon ControlNet~\cite{zhang2023adding}, enables pose-guided video generation.
We finetune WAN-Fun on our dataset using projected 2D hand skeletons as control inputs.
For track-based control, we include WAN-Move~\cite{chu2025wan} and MotionStream~\cite{shin2025motionstream}, two recent state-of-the-art methods.
For WAN-Move, we report results from both the original pre-trained model and a version trained on our dataset, denoted as WAN-Move$^*$.
As the official implementation of MotionStream is not publicly released, we re-implement its architecture and train it under the same settings on our dataset for a fair comparison.

\subsection{Quantitative Comparisons}
\begin{table*}[t]
    \centering
    \caption{\textbf{Quantitative comparisons.} We evaluate performance on Ego4D~\cite{grauman2022ego4d} and EgoDex~\cite{hoque2025egodex} datasets using visual quality metrics (FID, FVD, PSNR, SSIM) and hand control precision metrics (MPJPE, MPVPE) (\bestcap{Best}, \secondcap{second-best}, and \thirdcap{third-best}).}
    \vspace{-3mm}
    \label{tab:main_results}
    \resizebox{\textwidth}{!}{%
    \begin{tabular}{l cccccc cccccc}
        \toprule
        \multirow{3}{*}{\textbf{Method}} & \multicolumn{6}{c}{\textbf{Ego4D}} & \multicolumn{6}{c}{\textbf{EgoDex}} \\
        \cmidrule(lr){2-7} \cmidrule(lr){8-13}
         & \multicolumn{4}{c}{Visual Quality}
         & \multicolumn{2}{c}{3D Hand Accuracy}
         & \multicolumn{4}{c}{Visual Quality}
         & \multicolumn{2}{c}{3D Hand Accuracy} \\
         \cmidrule(lr){2-5} \cmidrule(lr){6-7}
         \cmidrule(lr){8-11} \cmidrule(lr){12-13}
         & FID $\downarrow$ & FVD $\downarrow$ & PSNR $\uparrow$ & SSIM $\uparrow$ 
         & MPJPE $\downarrow$ & MPVPE $\downarrow$
         & FID $\downarrow$ & FVD $\downarrow$ & PSNR $\uparrow$ & SSIM $\uparrow$ 
         & MPJPE $\downarrow$ & MPVPE $\downarrow$ \\
        \midrule
        Mask2IV(H)~\cite{li2025mask2iv}& 106.14 & 675.15 & 10.50 & 0.271 & 17.10 & 4.49 & 93.74 & 516.84 & 19.63 & 0.764 & 7.63 & 2.10\\
        Mask2IV(HO)~\cite{li2025mask2iv}& 107.85 & 637.77 & 10.50 & 0.277 & 16.46 & 5.16 & 96.63 & 520.77 & 19.57  & 0.765 & 7.64 & 2.16\\
        
        WAN-Fun~\cite{Wan2_1_Fun_Control2025} & $\phantom{1}$\second{77.39} & \second{303.54} & \second{15.68} & \second{0.436} & $\phantom{1}$\second{1.76} & \second{1.78} & \second{39.92} & \second{178.67} & \second{24.55} & \second{0.842} & \second{1.85} & \second{1.88} \\
        MotionStream~\cite{shin2025motionstream} & $\phantom{1}$\third{87.48} & \third{308.63}  & \third{14.65} & \third{0.400} & $\phantom{1}$9.99 & \third{3.28} & 61.39 & \third{288.90} & 21.50 & 0.790 & 5.76 & 2.11\\
        
        WAN-Move~\cite{chu2025wan} & $\phantom{1}$88.78 & 452.40 & 14.10 & 0.387 & $\phantom{1}$\third{9.11} & 4.35& 57.13 & 292.40 & 22.42  & 0.808 & \third{5.21} & 2.23\\
        WAN-Move*~\cite{chu2025wan} & $\phantom{1}$88.90 & 332.44  & 14.37 & 0.380 & $\phantom{1}$9.77 & 4.59 & \third{52.31} & 316.87 & \third{23.39} & \third{0.822} & 5.71 & \third{2.04} \\
        \midrule
        \textbf{Ours} & $\phantom{1}$\best{67.70} & \best{259.99}  & \best{15.86} 
        & \best{0.443} & $\phantom{1}$\best{1.42} & \best{1.70}& \best{39.62} & \best{174.73} & \best{24.74}
        & \best{0.846} & \best{1.80} & \best{1.82}\\
        \bottomrule
    \end{tabular}
    }
    \vspace{-7mm}
\end{table*}
\begin{wrapfigure}{R}{0.5\textwidth}
    \vspace{-0.7cm} 
    \centering
    \includegraphics[width=\linewidth]{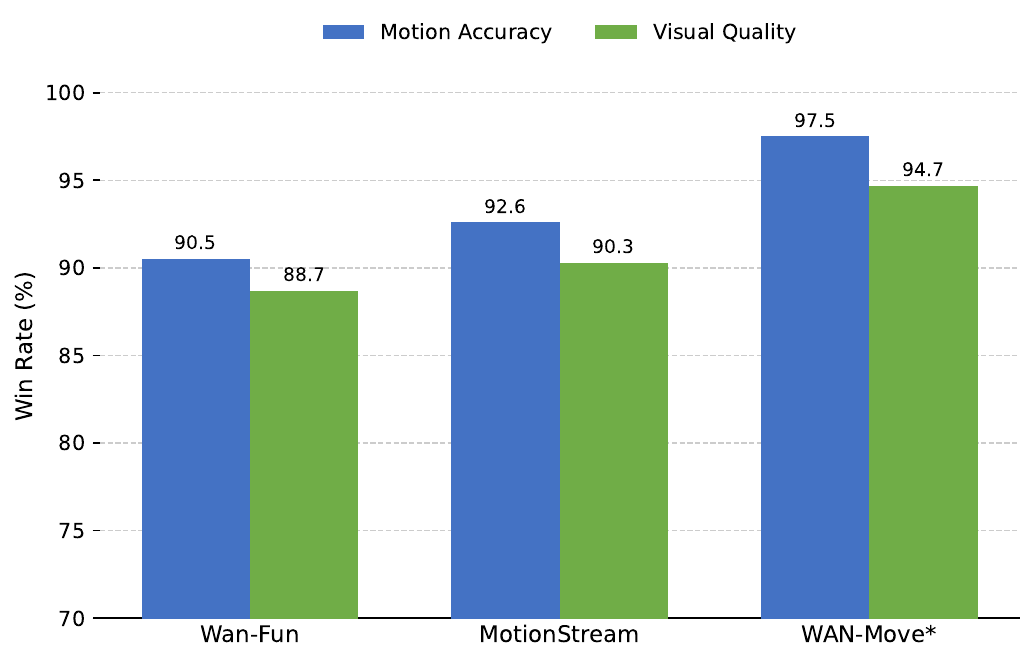}
    \vspace{-0.7cm} 
    \caption{The user study win rates.}
    \label{fig:2afc_win_rates}
    \vspace{-0.7cm} 
\end{wrapfigure}
Table \ref{tab:main_results} presents the quantitative comparisons with respect to visual fidelity and 3D hand trajectory accuracy.
Mask2IV, which attempts to constrain hand motion solely through 2D masks, suffers from significant visual clutter and depth ambiguity, yielding inferior results. 
WAN-Fun, a widely adopted pose-based baseline, is hindered by topological ambiguities inherent to 2D skeleton projections. 
Consequently, it falls behind our approach, showing a 14\% worse FVD and a 19\% higher MPJPE on the Ego4D dataset. 
Consistent performance gaps are observed on the EgoDex dataset, despite its relatively simpler interactions and cleaner backgrounds. 
When evaluated against recent state-of-the-art track-based methods MotionStream and WAN-Move$^*$ (training WAN-Move in egocentric data provides slight gains in video quality and negligible effects in hand accuracy), our explicit 3D joint-based approach achieves significantly better performance.
By explicitly resolving occlusion ambiguities and injecting 3D geometric priors, our method achieves at least a 16\% reduction in FVD (against MotionStream on Ego4D) and a 68\% reduction in MPJPE (against WAN-Move* on EgoDex).

Additionally, we present a Two-Alternative Forced Choice (2AFC) user study following \cite{chu2025wan}, as detailed in Figure \ref{fig:2afc_win_rates}. 
Based on evaluations from 30 participants across 30 diverse video sequences (15 sampled from each dataset), our method significantly outperforms current state-of-the-art competitors in both perceived video quality and human-evaluated motion accuracy.

\subsection{Qualitative Results}

\begin{figure}[t]
    \centering
    \includegraphics[width=\linewidth]{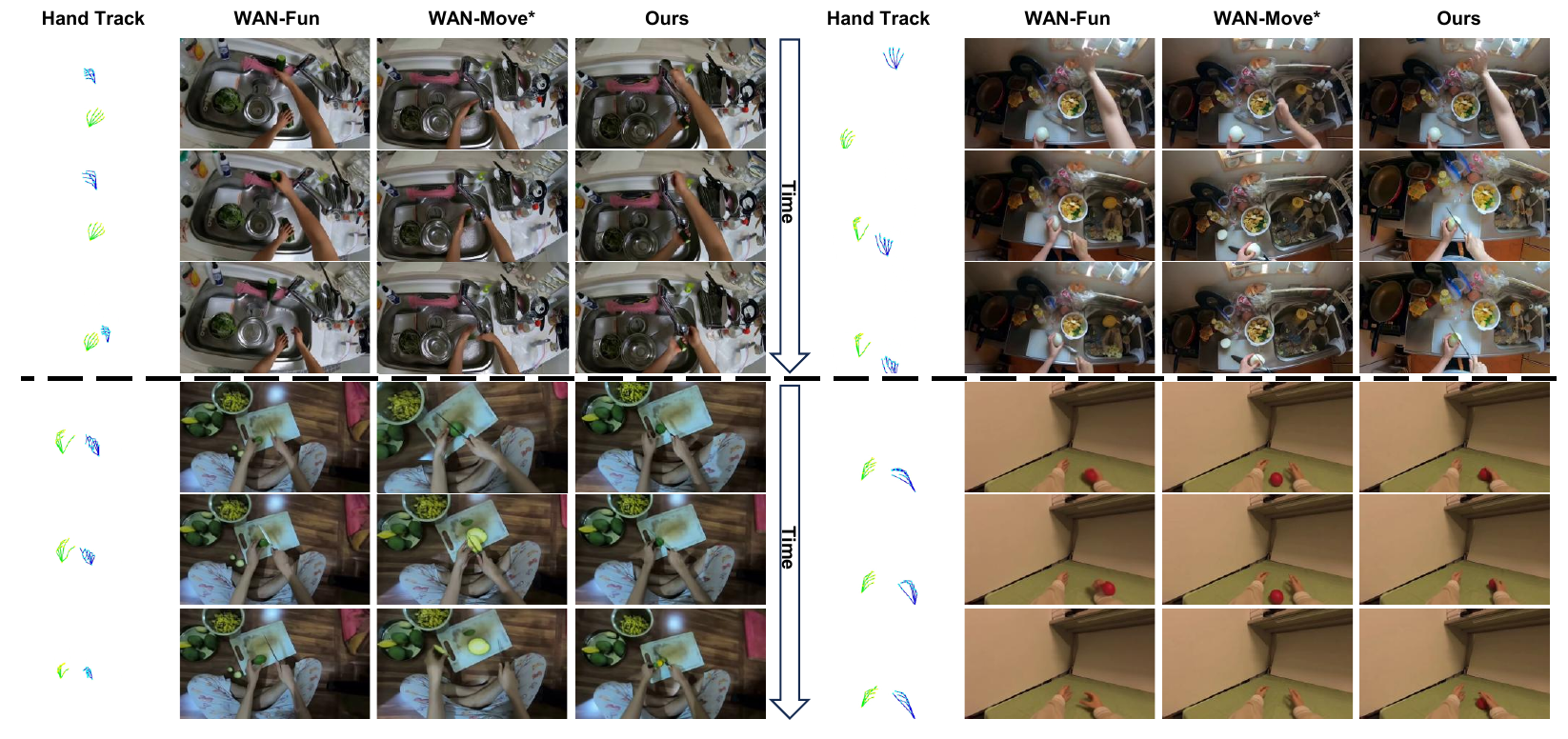}%
    \vspace{-3mm}
    \caption{\textbf{Qualitative comparisons.} 
    Compared with state-of-the-art WAN-Fun \cite{Wan2_1_Fun_Control2025} and WAN-Move$^*$ \cite{chu2025wan}, our method shows better video quality with accurate hand control.}
    \label{fig:vis}
    \vspace{-7mm}
\end{figure}

\noindent\textbf{Motion Control Comparisons.}
As shown in Figure~\ref{fig:vis}, given a reference hand track sequence, our method consistently outperforms the state-of-the-art pose-based WAN-Fun~\cite{Wan2_1_Fun_Control2025} and track-based WAN-Move*~\cite{chu2025wan} in both visual fidelity and motion control accuracy. 
WAN-Fun frequently generates physically implausible interactions due to its limited spatial understanding of fine-grained hand articulations. 
For example, in the top-left scenario, the model hallucinates a detached cucumber instead of accurately grasping the faucet to control the water flow. 
Similarly, in the bottom-left example, it fails to execute the mango-cutting action due to poor spatial grounding of the hand trajectory. 
WAN-Move* demonstrates generally weaker control authority, suffering from significant spatial drift because it relies solely on unstructured 2D point tracks. 
As highlighted in the bottom-right example, although it captures coarse 2D translation, it inherently neglects 3D hand rotation, a critical signal for complex interactions like manipulating a ball. 
In contrast, our 3D-joint-based conditioning ensures precise, geometrically consistent manipulations without structural hallucinations.

\begin{figure}[t]
    \centering
    \includegraphics[width=\linewidth]{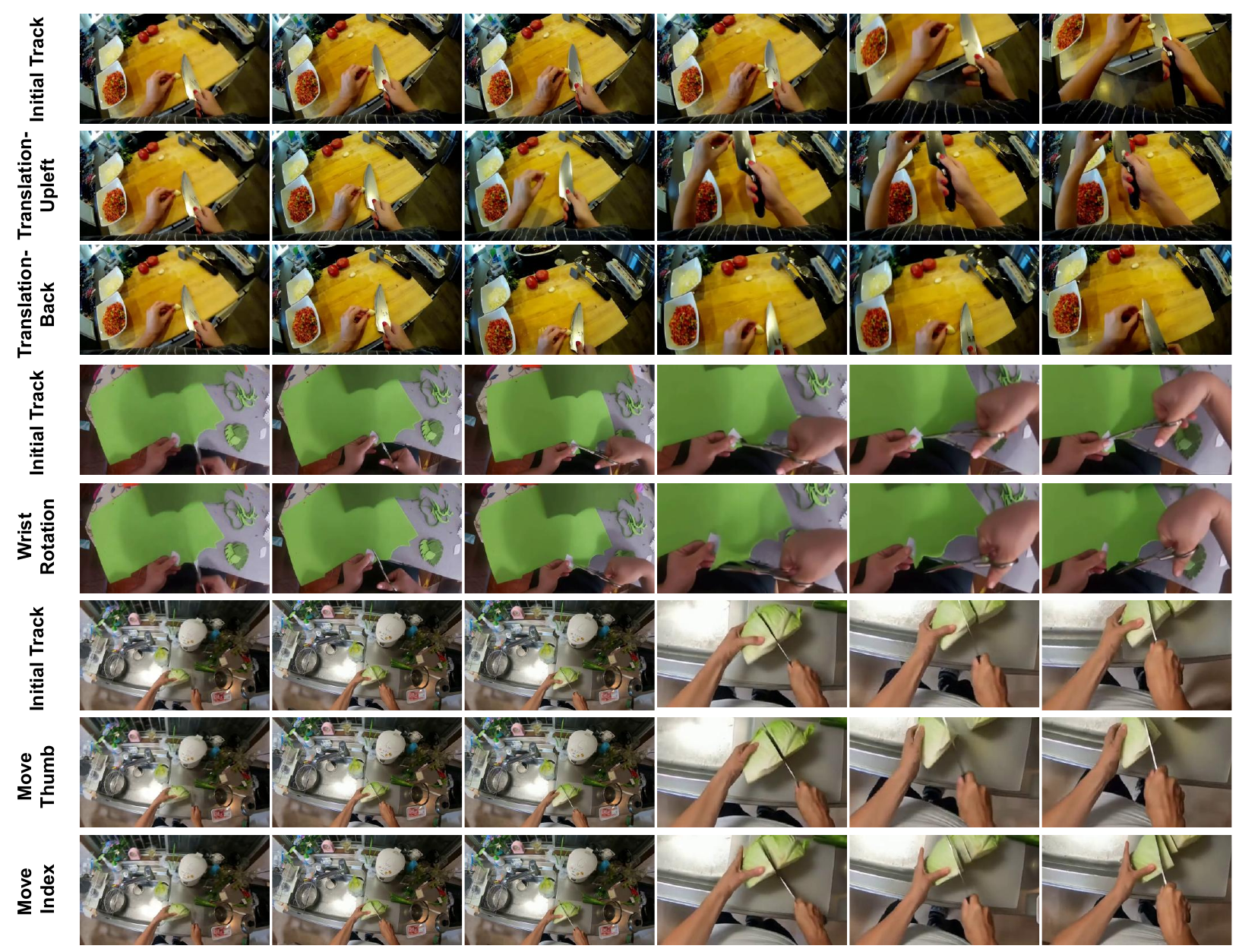}%
    \vspace{-3mm}
    \caption{\textbf{Interactive fine-grained hand control results.} 
    For the bottom five rows, the last three frames are zoomed in to show subtle hand changes.}
    \label{fig:vis_inter}
    \vspace{-5mm}
\end{figure}

\noindent\textbf{Interactive Fine-Grained Control.}\label{sec:inter_vis}
Enabled by the spatial grounding of the sparse 3D joints, our framework naturally supports interactive video editing. 
We develop an intuitive user interface that allows users to seamlessly modify hand trajectories by dragging single or multiple key joints in 3D space. 
As shown in Figure~\ref{fig:vis_inter}, our method achieves highly accurate, fine-grained control across various levels of movement, ranging from macro-adjustments, such as global hand translations and complex wrist rotations, to isolated micro-manipulations, such as moving a single digit (\textit{e.g.,} the thumb or index finger) while maintaining the structural integrity of the rest of the hand.

Furthermore, we provide additional results in the Supplementary Material demonstrating our control framework's cross-embodiment capabilities after fine-tuning on small-scale robotic dexterous hand data. 
A critical advantage of using 3D hand joints as the sole control modality is its inherent morphological invariance. 
Unlike rigid skeletal representations or implicit poses that strictly assume human-specific kinematic chains, bone lengths, and degrees of freedom, sparse 3D joints provide a purely spatial and kinematic-agnostic control signal ~\cite{qin2022dexmv, xin2026analyzing}.

\subsection{Ablation Studies}
\begin{table}[t]
    \centering
    \caption{\textbf{Ablation study of key components.} We sequentially analyze the contribution of Occlusion-Removed Context Aggregation (OCA), Occlusion-Aware Propagation (OP), and 3D Geometric Embeddings (3DGE) on Ego4D~\cite{grauman2022ego4d} and EgoDex~\cite{hoque2025egodex}.}
    \label{tab:ablation}
    \vspace{-3mm}
    \resizebox{\linewidth}{!}{%
    \begin{tabular}{ccc cccccc cccccc}
        \toprule
        \multicolumn{3}{c}{\textbf{Components}} & \multicolumn{6}{c}{\textbf{Ego4D}} & \multicolumn{6}{c}{\textbf{EgoDex}} \\
        \cmidrule(lr){1-3} \cmidrule(lr){4-9} \cmidrule(lr){10-15}
        \textbf{OCA} & \textbf{OP} & \textbf{3DGE} & FID $\downarrow$ & FVD $\downarrow$ & PSNR $\uparrow$ & SSIM $\uparrow$ & MPJPE $\downarrow$ & MPVPE $\downarrow$ & FID $\downarrow$ & FVD $\downarrow$ & PSNR $\uparrow$ & SSIM $\uparrow$ & MPJPE $\downarrow$ & MPVPE $\downarrow$ \\
        \midrule
        \checkmark & & & 80.14 & 305.49 & 14.97 & 0.405 & 2.75 & 2.81 & 41.05 & 198.86 & 23.62 & 0.822 & 1.94 & 1.99 \\
        \checkmark & \checkmark & & 76.08 & 300.99 & 15.28 & 0.417 & 2.10 & 2.27 & 40.27 & 190.01 & 24.18 & 0.839 & 1.89 & 1.86 \\
        \checkmark & \checkmark & \checkmark & \textbf{67.70} & \textbf{259.99} & \textbf{15.86}  & \textbf{0.443} & \textbf{1.42} & \textbf{1.70} & \textbf{39.62} & \textbf{174.73} & \textbf{24.74} & \textbf{0.846} & \textbf{1.80} & \textbf{1.82} \\
        \bottomrule
    \end{tabular}
    }
    \vspace{-7mm}
\end{table}

Table~\ref{tab:ablation} presents a quantitative analysis of the key components comprising our occlusion-aware 3D hand conditioning strategy. 
We establish our baseline by applying Occlusion-Removed Context Aggregation (OCA), which filters out contaminated visual features from occluded regions in the source frame. 
Applying this mechanism alone already yields substantial improvements in both visual quality and trajectory adherence compared to the recent state-of-the-art 2D track-based method~\cite{chu2025wan}.
Building upon OCA, the introduction of Occlusion-Aware Propagation (OP) addresses target occlusion ambiguities during dynamic interactions, achieving a further 5\% enhancement in FID and a 23\% reduction in MPJPE on the Ego4D dataset, with consistent performance gains observed on EgoDex.
Finally, we demonstrate that injecting explicit 3D geometric and semantic priors is vital to achieving true 3D-aware trajectory conditioning. 
3D Geometric Embeddings (3DGE) further boosts visual quality by 11\% (FID) and improves structural control accuracy by 32\% (MPJPE) on Ego4D. 

\subsection{Robotic Applications}

\begin{figure}[t]
    \centering
    \includegraphics[width=\linewidth]{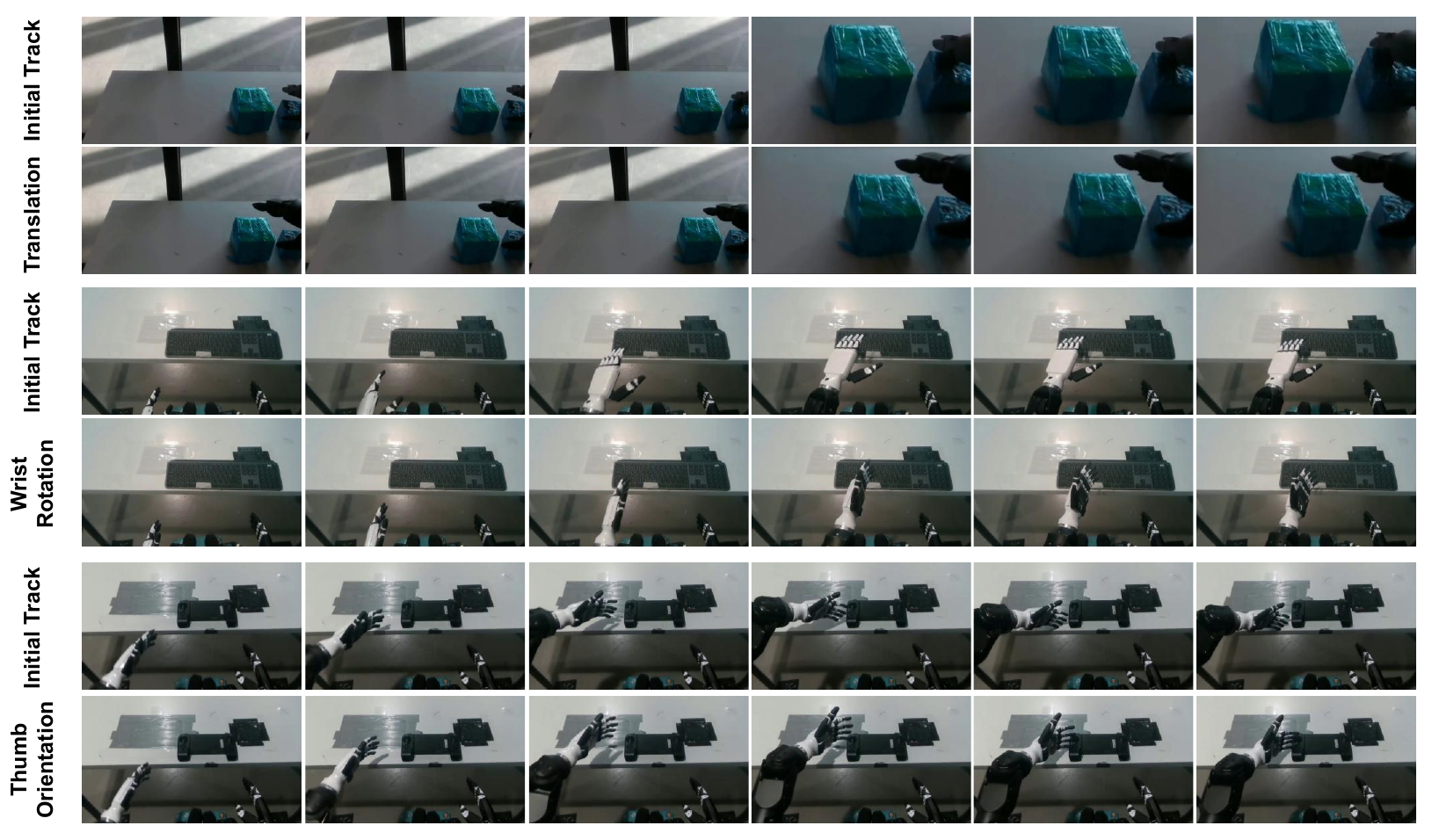}%
    \vspace{-3mm}
    \caption{\textbf{Interactive control results on diverse robotic hands.} 
    The last three frames in the first two rows are cropped and magnified for enhanced visualization.}
    \label{fig:supp_vis_inter}
    \vspace{-7mm}
\end{figure}

Because the morphological invariant sparse 3D hand joints align the spatial trajectories of both human and robotic hands \cite{qin2022dexmv, xin2026analyzing}, our framework successfully bridges the embodiment gap, effectively controlling robotic hands while requiring only minimal finetuning (Details in the Supplementary Material).
Figure~\ref{fig:supp_vis_inter} presents qualitative results of our interactive, fine-grained control applied to robotic scenarios. 
Crucially, these complex manipulations across entirely different robotic structures (Unitree G1-Dex3-1 and H1-Inspire provided in Humanoid Everyday~\cite{zhao2025humanoid} dataset) are achieved using a single, unified finetuned model. 

\section{Conclusion}
This paper proposes a novel, lightweight framework for high-fidelity, motion-controlled video generation in egocentric scenarios. 
By leveraging sparse 3D hand joints as explicit control signals, our approach overcomes the limitations of 2D tracks and implicit pose representations, which frequently fail to capture precise dexterous manipulations under mutual occlusion. 
We introduce an occlusion-aware 3D conditioning strategy that penalizes unreliable source signals and propagates features using a 3D-based weighting mechanism towards target frames. 
Coupled with the direct injection of 3D geometric embeddings, our method ensures robust structural consistency while minimally disrupting the generative priors of the pre-trained backbone. 
Additionally, we contribute a large-scale automated annotation pipeline for the Ego4D dataset, yielding over one million high-quality clips with accurate hand trajectories. 
Extensive experiments validate that our method achieves superior accuracy in controlling complex egocentric interactions compared to existing baselines.

\section*{Acknowledgements}
We would like to thank peers who helped on the development of the project: Zhenyu Zhao, Guanlong Jiao, Alexey Gavryushin, Jovana Videnovic and Nao Wu.
This work was supported 
(1) as part of the Swiss AI Initiative by a grant from the Swiss National Supercomputing Centre (CSCS) under project ID a03 on Alps; 
(2) by the Swiss National Science Foundation Advanced Grant 216260: Beyond Frozen Worlds: Capturing Functional 3D Digital Twins from the Real World;
(3) by European Union’s Horizon Europe research and innovation programme under grant agreement number 101214398 (ELLIOT), fully funded by Swiss State Secretariat for Education, Research and Innovation (SERI);
(4) by the ETH AI Center through an ETH AI Center doctoral fellowship to Botao Ye and Boqi Chen; 
(5) by the Max Planck ETH Center for Learning Systems (CLS) through a doctoral fellowship to Alexandros Delitzas.

%
%
\bibliographystyle{splncs04}
\bibliography{main}
\end{document}